\documentclass[a4paper,fleqn]{cas-sc}
\usepackage[numbers,sort&compress]{natbib}
\usepackage{mathtools}
\usepackage{multirow}
\usepackage{float}
\usepackage{subcaption}
\usepackage{bm}
\usepackage{threeparttable}
\usepackage{rotating} 
\usepackage{adjustbox}
\newcommand{\figref}[1]{Fig.~\ref{#1}}
\newcommand{\subsecref}[1]{Subsection~\ref{#1}}
\newcommand{\secref}[1]{Section~\ref{#1}}

\def\tsc#1{\csdef{#1}{\textsc{\lowercase{#1}}\xspace}}
\tsc{WGM}
\tsc{QE}
\begin{document}
\let\WriteBookmarks\relax
\def\floatpagepagefraction{1}
\def\textpagefraction{.001}

\title [mode = title]{Motion Estimation for Multi-Object Tracking using KalmanNet with Semantic-Independent Encoding}  

\tnotemark[1] 

\tnotetext[1]{This research did not receive any specific grant from funding agencies in the public, commercial, or not-for-profit sectors.}

\author[1]{Jian Song}[
      orcid=0009-0009-8891-5320]
\ead{songj9507@outlook.com}
\credit{Conceptualization, Writing - Original draft preparation, Methodology, Software}

\author[1]{Wei Mei}[
  orcid=0000-0002-5736-189X]
\cormark[1]
\ead{meiwei@sina.com}
\credit{Conceptualization, Methodology, Validation, Writing - Review \& Editing, Supervision, Project administration}

\author[1,2]{Yunfeng Xu}[
  orcid=0000-0002-2674-3721]
\ead{hbkd_xyf@hebust.edu.cn}
\credit{Conceptualization, Methodology, Visualization, Investigation}

\author[1]{Qiang Fu}[
  orcid=0000-0002-3831-9856]
\ead{Fu_Qiang@aeu.edu.cn}
\credit{Data Curation, Resources}

\author[3]{Renke Kou}[
  orcid=0000-0001-5893-3127]
\ead{krkoptics@163.com}
\credit{Software, Data Curation, Resources, Investigation}

\author[1]{Lina Bu}[
  orcid=0009-0005-8391-3423]
\ead{bulina90@aeu.edu.cn}
\credit{Visualization, Formal analysis}

\author[1]{Yucheng Long}[
  orcid=0009-0000-6328-5494]
\ead{longyc0010@163.com}
\credit{Visualization, Validation}

\affiliation[1]{organization={Army Engineering University},%Department and Organization
city={Shijiazhuang},
postcode={050003}, 
state={Heibei},
country={China}}

\affiliation[2]{organization={Heibei University of Science and Technology},
city={Shijiazhuang},
postcode={050003}, 
state={Heibei},
country={China}}

\affiliation[3]{organization={Air Force Engineering University},
            city={Xi'an},
            postcode={710000}, 
            state={Shanxi},
            country={China}}

\cortext[1]{Corresponding author}

\begin{abstract}
  Motion estimation is a crucial component in multi-object tracking (MOT).
  It predicts the trajectory of objects by analyzing the changes in their positions in consecutive frames of images, reducing tracking failures and identity switches. 
  The Kalman filter (KF) based on the linear constant-velocity model is one of the most commonly used methods in MOT. 
  However, it may yield unsatisfactory results when KF's parameters are mismatched and objects move in non-stationary.
  In this work, we utilize the learning-aided filter to handle the motion estimation of MOT.
  In particular, we propose a novel method named Semantic-Independent KalmanNet (SIKNet), which encodes the state vector (the input feature) using a Semantic-Independent Encoder (SIE) by two steps. 
  First, the SIE uses a 1D convolution with a kernel size of 1, which convolves along the dimension of homogeneous-semantic elements across different state vectors to encode independent semantic information. 
  Then it employs a fully-connected layer and a nonlinear activation layer to encode nonlinear and cross-dependency information between heterogeneous-semantic elements.
  To independently evaluate the performance of the motion estimation module in MOT, we constructed a large-scale semi-simulated dataset from several open-source MOT datasets. 
  Experimental results demonstrate that the proposed SIKNet outperforms the traditional KF and achieves superior robustness and accuracy than existing learning-aided filters.
  The code is available at (\url{https://github.com/SongJgit/filternet} and \url{https://github.com/SongJgit/TBDTracker}).
\end{abstract}

% Keywords

\maketitle

\section{Introduction}
\label{sec1}
%% Labels are used to cross-reference an item using \ref command.
Online and real-time multi-object tracking (MOT) plays a fundamental role in various applications, such as autonomous driving \cite{2020weng3dmultiobjecttracking,2021chiuprobabilistic3dmultimodal} and motion analysis \cite{2022cioppasoccernettrackingmultipleobject}. 
Benefiting from the development of deep learning-based detectors \cite{2015girshickfastrcnn, 2017renfasterrcnnrealtime,2020ciaparronedeeplearningvideo,2021geyoloxexceedingyolo}, MOT methods based on tracking-by-detection (TBD) paradigm have attracted considerable attention.
The basic idea behind these methods is to track the detection bounding-boxes  (bbox) provided by the detector \cite{2016bewleysimpleonlinerealtime,2017wojkesimpleonlinerealtime, 2021zhangfairmotfairnessdetection,2021dugiaotrackercomprehensiveframework, 2022zhangbytetrackmultiobjecttracking, 2023yanghardtrackobjects, 2024qimultipleobjecttracking,2024wangdeeplearningmultimodal}.
Most trackers that follow the TBD paradigm can be divided into two components: Motion Estimation Module (MEM) and Data Association Module (DAM). 
Specifically, the DAM associates the detections with the motion predictions based on a certain metric, e.g., intersection-over-union (IoU), to update the trajectory \cite{1955kuhnhungarianmethodassignment}. 
This makes the tracking performance highly dependent on the predictions of the MEM. 
Most of these state-of-the-art (SOTA) trackers adopt the model-based Kalman filter (KF), which typically uses on a linear constant-velocity (CV) model \cite{2024wangdeeplearningmultimodal}.
However, the linear CV assumption is often violated in practical scenarios, as noise parameters and object motion patterns are rarely fully known.
Detectors with varying performances and moving cameras often introduce additional uncertainties, leading to model mismatch and significant degradation of motion estimation.
This results in a small IoU and incorrect match between the predicted and detected results, triggering tracking failures \cite{2023yanghardtrackobjects, 2022zhudetectiontrackingmeet, 2024qimultipleobjecttracking, 2024yiucmctrackmultiobjecttracking, 2024lvdiffmotrealtimediffusionbased, 2025wangpreformermottransformerbased}.
This might explain why algorithms performing well on the pedestrian-dominant MOT17/MOT20 datasets \cite{2016milanmot16benchmarkmultiobject,2020dendorfermot20benchmarkmulti} suffer significant performance degradation on datasets such as SoccerNet \cite{2022cioppasoccernettrackingmultipleobject,2021cioppacameracalibrationplayer} and DanceTrack \cite{2022sundancetrackmultiobjecttracking}, whose object motion patterns are far more complex and difficult to be described by linear CV model.

Some researches have attempted to solve the model mismatch problem using variants of the KF.
For example, Patil \cite{2019patilh3ddatasetfullsurround} applied the unscented Kalman filter to estimate the object's linear and angular velocities.
Liu \cite{2021liuimmenabledadaptive3d} and Qi \cite{2024qimultipleobjecttracking} attempted to overcome the decline in the estimation performance of the KF when there is model mismatch by introducing the interacting multiple model.
On the other side, the above methods usually require statistical information about the datasets to configure the parameters of the filter, and the same parameters will be used for all objects, making it difficult to adapt to different object categories (e.g., ``Pedestrian'' and ``Dancer'').

In this paper, we aim to explore KalmanNet—a type of learning-aided Kalman filtering (LAKF) model \cite{2022shlezingermodelbaseddeeplearning,2023shlezingermodelbaseddeeplearning, 2021revachkalmannetdatadrivenkalman,2022revachkalmannetneuralnetwork,2023choisplitkalmannetrobustmodelbased} as alternatives for motion estimation in MOT.
In addition, we propose an improved KalmanNet, Semantic-Independent KalmanNet (SIKNet).
The learning-aided filters exploit recurrent neural networks (RNNs) instead of analytic models to adaptively learn Kalman gain (KG) from input features, such as measurement residuals.
This allows the motion estimation algorithm to better adapt to the nonlinear motion of various objects, resulting in more accurate motion predictions even in scenarios where the object motion is non-stationary, and the camera motion is unknown.
To improve the robustness of KNet, we incorporated a Semantic-Independent Encoder (SIE) for input-feature embedding. 
This encoder is used to extract independent semantic information from homogeneous-semantic elements of different state vectors and the cross-dependencies between heterogeneous-semantic elements. 
The SIE can effectively reduce the adverse impact on the stability of the network's training caused by heterogeneous-semantic elements, which in general have a large different scale.

To evaluate the MEM, we systematically constructed a semi-simulated dataset by integrating several real-world open-source datasets \cite{2016milanmot16benchmarkmultiobject,2020dendorfermot20benchmarkmulti,2022cioppasoccernettrackingmultipleobject,2022sundancetrackmultiobjecttracking}. 
As we know, the detection results in existing MOT datasets have not undergone data association matching, which means they cannot be used directly for independent MEM evaluation. 
The semi-simulated dataset can help us make such an independent analysis of the MEM in MOT task. 
Experimental results show that the SIKNet with the Semantic-Independent Encoder is more stable in training, and exhibits more robust and accurate estimation performance compared to model-based KF and existing learning-aided filters. 

\textit{Our main contributions are as follows:}

1) We propose a novel Semantic-Independent KalmanNet to improve the accuracy and robustness of motion estimation in the MOT task.

2) We construct a semi-simulated dataset for independent evaluation of the motion estimation module in MOT.

3) We develop an open-source framework, FilterNet, which allows researchers to reproduce and compare KF, KNet, Split-KalmanNet(SKNet) and SIKNet easily in a unified way.
 
The rest of this paper is organized as follows:
\secref{sec2} introduces the task of motion estimation in MOT using KF as well as a short review of KNet.
\secref{sec3} describes the semi-simulated datasets.
\secref{sec4} presents the developed Semantic-Independent KalmanNet.
\secref{sec5} presents the experimental results.
\secref{sec6} concludes the paper.

\section{Preliminaries}\label{sec2}
\subsection{Kalman motion estimation in multi-object tracking} \label{subsec2.1}

\subsubsection{System model}
Consider a discrete-time state space model represented via

\begin{align}
    \bm{x}_{t}&=\bm{f}(\bm{x}_{t-1})+\bm{e}_t,&\bm{x}_t&\in\mathbb{R}^m, \label{eq:systemmodel1}\\
    {\boldsymbol{y}}_{t}&=\bm{h}(\bm{x}_{t})+\bm{v}_t,&\bm{y}_t&\in\mathbb{R}^n,\label{eq:systemmodel2}
\end{align}
where $\bm{x}_t$ and $\bm{y}_t$ are state and measurement vectors at time $t$. 
$\bm{e}_{t}$ and $\bm{v}_{t}$ are Gaussian random vectors that have zero-mean and unknown covariance matrices $\bm{Q}_{t}=\mathbb{E}[\bm{e}_t\bm{e}_t^\top]$ and $\bm{R}_{t}=\mathbb{E}[\bm{v}_t\bm{v}_t^\top]$. 
$\bm{f}(\cdot)$ and $\bm{h}(\cdot)$ are the state transition and measurement functions.

\subsubsection{State mode}
In the motion estimation modules of most MOT algorithms\cite{2016bewleysimpleonlinerealtime,2017wojkesimpleonlinerealtime,2020weng3dmultiobjecttracking,2021chiuprobabilistic3dmultimodal,2023yanghardtrackobjects,2022zhangbytetrackmultiobjecttracking,2023dustrongsortmakedeepsort,2024qimultipleobjecttracking,2023caoobservationcentricsortrethinking,2021zhangfairmotfairnessdetection}, the state of each object is modeled as:
\begin{equation} \label{eq:state}
\bm{x}_t=[cx,\dot{cx}, cy, \dot{cy}, a, \dot{a},h, \dot{h}]^\top,
\end{equation}
where $[cx, cy, a, h]$ is used to represent the center position, aspect ratio, and height of an object in an image, it stands for the object's bounding-box (bbox).
$[\dot{cx}, \dot{cy}, \dot{a}, \dot{h}]$ denote the velocities of $cx$, $cy$, $a$, and $h$.

Correspondingly, the measurement $\bm{y}_t$ from the detector is related to the object bbox $[cx, cy, a, h]^\top$.
For convenience, we refer to this state representation method as the \textit{XYAH} mode.

Another commonly used state mode is \cite{2022aharonbotsortrobustassociations,2024yiucmctrackmultiobjecttracking}:
\begin{equation} \label{eq:state2}
\bm{x}_t=[cx,\dot{cx}, cy, \dot{cy}, w, \dot{w},h, \dot{h}]^\top,
\end{equation}
where $[w, h]$ are the width and height of the bounding-box, and $[\dot{w}, \dot{h}]$ are the corresponding velocities. 
Correspondingly, the measurement $\bm{y}_t$ is related to the object bbox $[cx, cy, w, h]^\top$, which is hence called \textit{XYWH} mode.

\subsubsection{Prediction stage}
Following the model-based KF, we can obtain the formula for the state prediction stage as follows:
\begin{subequations}
    \label{eq:kfpred}
\begin{align}
     &\bm{\hat{x}}_{t|t-1}=\bm{f}(\bm{\hat{x}}_{t-1|t-1}), \label{eq:kfpred1}\\
     &\bm{{P}}_{t|t-1}=\bm{F}_{t-1}\bm{{P}}_{t-1|t-1}\bm{F}_{t-1}^\top+\bm{Q}_{t}, \label{eq:kfpred2}
\end{align}
\end{subequations}
where $\bm{\hat{x}}_{t-1|t-1}$ is the estimated mean of the true state $\bm{x}$ at time $t-1$, and $\bm{\hat{x}}_{t|t-1}$ is the predicted mean at time $t$.
The matrix $\bm{F}_{t-1}=\frac{\partial\bm{f}}{\partial\bm{x}}|_{\bm{x}=\bm{\hat{x}}_{t-1|t-1}}$ is the Jacobian of $\bm{f}(\cdot)$ evaluated at $\bm{\hat{x}}_{t-1|t-1}$.
The matrix $\bm{{P}}_{t-1|t-1}$ and $\bm{{P}}_{t|t-1}$ are updated and predicted state covariances, respectively. 
$\bm{Q}_{t}$ is the process noise covariance matrix.

In MOT, a linear CV model is commonly used to approximate the inter-frame displacement of each object, and it can be expressed as
\begin{equation}
        \bm{f}(\bm{\hat{x}}_{t-1|t-1})=\bm{F}\bm{\hat{x}}_{t-1|t-1} \\
\end{equation}
where, 
\begin{equation}
    \label{F}
    \begin{aligned}
    \bm{F}=\mathrm{diag}[\bm{F}_{cv},\bm{F}_{cv},\bm{F}_{cv},\bm{F}_{cv}], \quad \bm{F}_{cv}=\begin{bmatrix}1&\Delta{t}\\0&1\end{bmatrix},
    \end{aligned}
\end{equation}
where $\Delta{t}$ is time interval. 
And, the process noise covariance matrix $\bm{Q}_{t}$ can be set up by using
\begin{equation}
    \label{xyahnoiseq}
    \begin{aligned}
    \bm{Q}_t=\mathrm{diag}[\bm{q}^2\circ\bm{q}_{d}^2],
    \end{aligned}
\end{equation}
where $\circ$ symbol means the element-wise multiplication. 
$\bm{q}$ and $\bm{q}_{d}$ for the \textit{XYAH} mode are defined as
\begin{equation}
    \begin{aligned}
        &\bm{q}=[\alpha_p, \alpha_v, \alpha_p, \alpha_v, 0.01, 0.00001, \alpha_p, \alpha_v], \\
        &\bm{q}_d=[h, h, h, h, 1, 1, h,h].
    \end{aligned}
    \end{equation}
Here, $\alpha_p$ and $\alpha_v$ serve as noise factors (hyperparameters) for position and velocity, respectively, 
 and $h$ is the height in the $\bm{\hat{x}}_{t-1|t-1}$.
For \textit{XYWH} mode, these parameters become
    \begin{equation}
    \begin{aligned}
        &\bm{q}=[\alpha_p, \alpha_v, \alpha_p, \alpha_v, \alpha_p, \alpha_v, \alpha_p, \alpha_v], \\
        &\bm{q}_d=[w, h, w, h, w, h, w,h].
    \end{aligned}
    \end{equation}
Here, $w$ is the width in the $\bm{\hat{x}}_{t-1|t-1}$.

\subsubsection{Update stage}
When the detected bounding-box is associated with an object, it will be used to update the object state:
\begin{subequations} \label{eq: kfupdate}
    \begin{align}
        &\bm{\hat{y}}_{t|t-1}=\bm{H}(\bm{\hat{x}}_{t|t-1}), \label{eq:kfupdate1}\\
        &\bm{S}_{t}=\bm{H}_{t}\bm{{P}}_{t|t-1}\bm{H}_{t}^\top+\bm{R}_{t}, \label{eq:kfupdate2}\\
        &\bm{K}_{t}=\bm{P}_{t|t-1}\bm{H}_{t}^\top\bm{S}_{t}^{-1}, \label{eq:kfupdate3}\\
        &\bm{\hat{x}}_{t|t}=\bm{\hat{x}}_{t|t-1}+\bm{K}_{t}(\bm{y}_{t}-\bm{\hat{y}}_{t|t-1}), \label{eq:kfupdate4}\\  
        &\bm{P}_{t|t}=(\bm{I}-\bm{K}_{t}\bm{H}_t)\bm{{P}}_{t|t-1},\label{eq:kfupdate5}
    \end{align}
\end{subequations}
where $\bm{\hat{y}}_{t|t-1}$ is the predicted  measurement at time $t-1$.
$\bm{H}_t=\frac{\partial\bm{h}}{\partial\bm{x}}|_{\bm{x}=\bm{\hat{x}}_{t|t-1}}$ is the Jacobian of $\bm{h}(\cdot)$ evaluated at $\bm{\hat{x}}_{t|t-1}$.
The matrix $\bm{S}_{t}$ is the innovation covariance, and $\bm{K}_{t}$ is the KG.

In MOT, $\bm{h}(\cdot)$ is also described as a linear model, that is,
\begin{equation}
    \bm{h}\left(\bm{\hat{x}}_{t|t-1}\right)=\bm{H}\bm{\hat{x}}_{t|t-1}
\end{equation}
where
\begin{equation}
    \label{H}
    \bm{H}=\mathrm{diag}[\bm{H}_{1},\bm{H}_{1},\bm{H}_{1},\bm{H}_{1}], \quad \bm{H}_{1}=\begin{bmatrix}1&0\end{bmatrix}.
\end{equation}
And, the measurement noise covariance matrix $\bm{R}_{t}$ is defined as 
\begin{equation}
    \label{xyahnoiser}
    \begin{aligned}
    \bm{R}_t=\mathrm{diag}[\bm{r}^2\circ\bm{r}_{d}^2],
    \end{aligned}
\end{equation}
where $\bm{r}$ and $\bm{r}_{d}$ for the \textit{XYAH} mode are defined as
\begin{equation}
    \label{xyahnoiserd}
    \begin{aligned}
        \bm{r}&=[\alpha_p, \alpha_p, 0.1, \alpha_p], &&\bm{r}_d=[h, h, 1, h].
    \end{aligned}
\end{equation}
For the \textit{XYWH} mode, these parameters become
\begin{equation}
    \begin{aligned}
        \bm{r}&=[\alpha_p, \alpha_p, \alpha_p , \alpha_p], &&\bm{r}_d=[w, h, w, h].
    \end{aligned}
\end{equation}

\subsubsection{Problem formulation}
The task of motion estimation in MOT is to predict the current object bbox based on the previous one and then to update it to obtain a more accurate bbox according to the associated detected results (measurements).
A more accurate bbox estimation is expected to get close to ground-truth (GT), which will improve the accuracy of data association and hence the overall performance of MOT. 
In this paper, the IoU between the estimated bbox $\bm{\hat{x}}_{t}^{b}$ and the GT $\bm{x}_{t}^{b}$, as given in Eq. (\ref{eq:iou}), is selected as the basic metric for motion estimation.
\begin{equation} \label{eq:iou}
\mathrm{IoU} = \frac{|\bm{\hat{x}}_{t}^{b}\cap \bm{x}_{t}^{b}|}{|\bm{\hat{x}}_{t}^{b}\cup  \bm{x}_{t}^{b}|}
\end{equation}
where $\bm{x}_t^{b}=[cx, cy, w, h]^{\top}$ or $[cx, cy, a, h]^{\top}$.
In \secref{sec5}, the basic metric IoU will be used to form the statistic metric of recall (Re) for evaluating the estimation performance. 
\subsection{KalmanNet} \label{subsec:kalmannet}
KalmanNet (KNet) \cite{2021revachkalmannetdatadrivenkalman,2022revachkalmannetneuralnetwork} has a similar overall structure to the model-based KF.
The only difference is that KNet exploits deep neural networks (DNNs) instead of analytical models to learn the KG using input features, thus gaining the ability to overcome model mismatch and nonlinearity. 
In addition, such learning-aided filters do not require prior statistical knowledge of the estimation problem, such as $\bm{Q}_{t}$ and $\bm{R}_{t}$.

Eqs. (\ref{eq:kfpred2}) and (\ref{eq:kfupdate2})-(\ref{eq:kfupdate5}) in the model-based KF are replaced by
\begin{equation} \label{eq5}
    \bm{\hat{x}}_{t|t}=\bm{\hat{x}}_{t|t-1}+\bm{K}_t(\theta)(\bm{y}_t-\bm{\hat{y}}_{t|t-1}),
\end{equation}
where $\bm{K}_t(\theta)$ is the learned KG using a DNN with trainable parameters $\theta$.
At each time step $t$, it uses the following features as inputs to learn the KG.
\begin{equation}
\begin{aligned} \label{eq:knetfeat}
    &\Delta\bm{\tilde{x}}_t=\bm{\hat{x}}_{t|t}-\bm{\hat{x}}_{t-1|t-1}, &&\Delta\bm{\hat{x}}_t=\bm{\hat{x}}_{t|t}-\bm{\hat{x}}_{t|t-1}, \\
    &\Delta\bm{\tilde{y}}_t=\bm{y}_t-\bm{y}_{t-1},&&\Delta\bm{y}_t=\bm{y}_t-\bm{\hat{y}}_{t|t-1}. 
\end{aligned}
\end{equation}
The state difference features available at time $t$ are $\Delta{\bm{\tilde{x}}_{t-1}}$ and $\Delta{\bm{\hat{x}}_{t-1}}$, respectively.

KNet underutilize the model-based (MB) structure when computing the KG, leading to instability of network training. 
To solve this problem, the Split-KalmanNet (SKNet) \cite{2023choisplitkalmannetrobustmodelbased} uses two parallel RNNs to track the covariance matrices $\bm{{P}}_{t|t-1}$ and $\bm{S}^{-1}_t$, respectively, which allows computing the KG via Eq. (\ref{eq:kfupdate3}). 
Song \cite{2024songpracticalimplementationkalmannet} proposed a sliding-window method to improve training robustness of the KNet and SKNet, which makes use of the first-order Markov property of the KF. 
However, all these methods still suffer from a disgusting problem of training instability, which to our understanding stems from the huge scale-difference of various heterogeneous-semantic elements of the input features.

\section{Semi-simulated datasets} \label{sec3}
The large-scale semi-simulated datasets are built upon the commonly used open-source MOT datasets.

\subsection{Base datasets} \label{subsec3.1}
{\bf MOT17 \cite{2016milanmot16benchmarkmultiobject} and MOT20 \cite{2020dendorfermot20benchmarkmulti}}.
These are large datasets used to evaluate the performance of tracking algorithms in traditional scenarios, with the main object category being ``Pedestrian''.
MOT17 is dominated by close-range video captured by dynamic cameras, while MOT20 focuses on dense pedestrian scenarios with wide-angle lens from near-static cameras, resulting in smaller object sizes. 
Notably, most objects in both datasets exhibit regular motion patterns that can be effectively approximated by linear CV motion models when camera motion is ignored \cite{2022sundancetrackmultiobjecttracking}.

{\bf SoccerNet \cite{2022cioppasoccernettrackingmultipleobject}}.
This dataset constitutes a large-scale benchmark for football video comprehension, containing annotations of ``Player'' and their movement trajectories throughout matches.
Most footage is captured from a fixed wide-angle lens, with ``Player'' displaying highly irregular and non-stationary motion regardless of camera motion.

{\bf DanceTrack \cite{2022sundancetrackmultiobjecttracking}}.
This dataset serves as a large-scale benchmark for dance video comprehension, containing annotations of ``Dancer'' and their trajectories. Most videos are captured at a fixed location near the stage. 
Compared with MOT17, MOT20, and SoccerNet datasets, ``Dancer'' movements exhibit higher irregularity and more significant bounding-box size variations, primarily due to the camera's closer proximity to the subjects.

{\bf{Analysis of Object Motion Pattern}}. 
Referring to \cite{2022sundancetrackmultiobjecttracking}, we use the ``IoU on adjacent frames'' metric to analyze the motion patterns of the main object categories in the dataset.
A low IoU suggests significant object motion between adjacent frames.
Given a video with \textit{N} targets and \textit{T} frames, the average IoU (AIoU) can be calculated by:
\begin{equation}
    \mathrm{AIoU}=\frac{1}{N(T-1)}\sum_{i}^{N}\sum_{t=1}^{T-1}\text{IoU}(b_t^i,b_{t+1}^i),
\end{equation}
where $\text{IoU}(\cdot)$ is the intersection-over-union of the bounding-box $\textit{b}$ on the adjacent frames for a certain object $i$.

\begin{figure}
     \centering
     \includegraphics[width=0.5\textwidth]{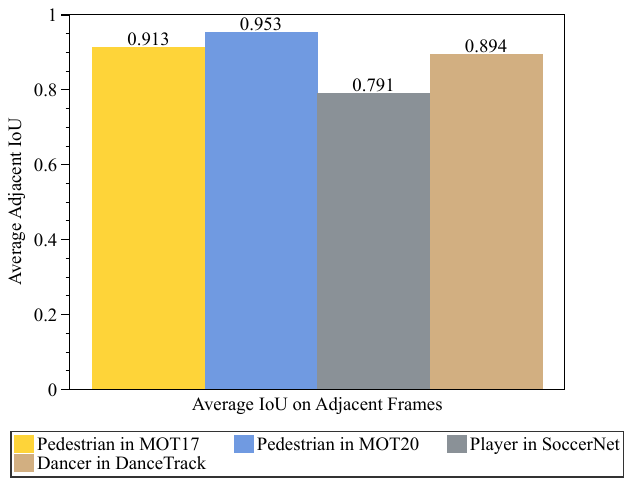}
     \caption{IoU on adjacent frames. Compared to MOT17 and MOT20, SoccerNet and DanceTrack have lower IoU values, which indicates that the objects in these datasets have more complex motion patterns.}
     \label{fig:analysis}
 \end{figure}
As shown in \figref{fig:analysis}, MOT20 has a higher ``Pedestrian'' IoU than MOT17 because MOT20 was captured by a stationary camera with a wide-angle lens, which reduces bounding-box size variations.
MOT17 is shot at close range and produces larger sized object bounding-boxes where limb movement patterns (e.g., shopping bag oscillations) significantly amplify adjacent frame bounding-box variations, resulting in depressed IoU metrics.
Notably, despite adopting the same fixed shot method as MOT20, SoccerNet and DanceTrack have relatively low IoU values due to the drastic changes in the object motion patterns.
In addition, SoccerNet's IoU is significantly lower than DanceTrack's because the ``Player'' have larger ranges of motion and higher velocities, resulting in larger inter-frame displacements and less bounding-box overlap.
MOT17 and MOT20 have relatively regular and stationary motion patterns, which are fit for evaluating the algorithm's performance in basic scenario.
In contrast, the complex motion patterns of SoccerNet and DanceTrack can be employed to test the algorithm's adaptability to model-mismatch and non-linearity.

\subsection{Semi-simulated datasets}\label{subsec3.2}
Existing MOT datasets, including the above four Datasets, provide the detected measurements and GT trajectory, which can be used for evaluating the tracker (including the motion estimation and data association modules) or the whole tracking framework (including both the detector and tracker). 
The measurements of these datasets have not been matched through data association; therefore they cannot be directly used for MEM evaluation. 
To address this problem, a method for generating semi-simulated datasets is proposed. 

The measurement sequence is to be generated only in the mode of \textit{XYAH} and will be transformed into \textit{XYWH} mode. 
The \textit{XYAH} and converted \textit{XYWH} sequences will then be used for the subsequent task of motion estimation.

We generated a semi-simulated dataset based on the \textit{XYAH} mode. 
Let $\bm{x}^{b}=[\bm{x}_1^{b},\bm{x}_2^{b} \ldots,\bm{x}_{T}^{b}]$ be a GT bbox sequence of length $T$ where $\bm{x}_t^{b}=[cx,cy,a,h]^{\top}$. 
For each time step $t$, the corresponding element $\bm{y}_{t}^b$ in the simulated bbox sequence $\bm{y}^{b}=[\bm{y}_1^{b},\bm{y}_2^{b}, \ldots,\bm{y}_{T}^{b}]$ can be obtained by:
\begin{equation} 
    \bm{y}_{t}^{b}= \bm{x}_t^{b} + \bm{v}_{t}, ~ \bm{v}_{t}\sim\mathcal{N}(\mathbf{0},\bm{R}_{t}).
    \label{eq:noise track}
\end{equation}
where the measurement noise covariance matrix $\bm{R}_{t}$ are defined in Eqs. (\ref{xyahnoiser}) and (\ref{xyahnoiserd}).

\begin{figure}[t]
     \centering
     \subfloat[The sequence are generated using $\alpha_p =0.05$]{%
         \includegraphics[width=\textwidth]{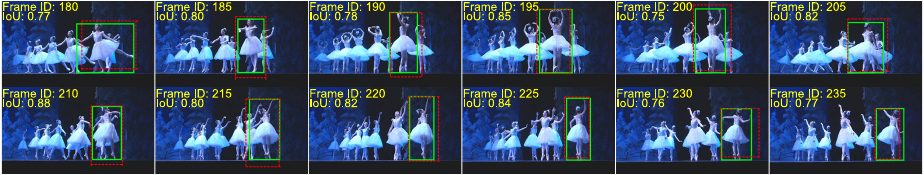}
     }
     \hfill
     \subfloat[The sequence are generated using $\alpha_p =0.4$]{%
         \includegraphics[width=\textwidth]{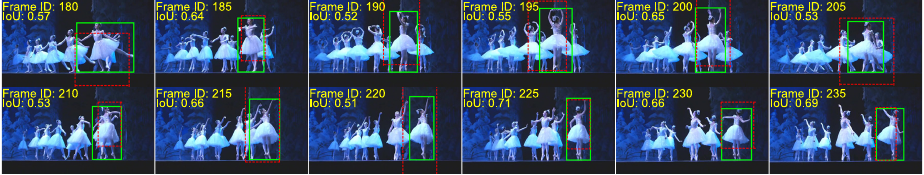}
     }
     \caption{Selected semi-simulated sequences on  DanceTrack video. Green solid lines denote ground-truth bounding-box, and the red dashed lines represent the noisy bounding-box $\bm{y}_{t}$ derived from Eq. (\protect\ref{eq:noise track}), and the IoU metric quantifies their spatial overlap.} 
     \label{fig:trackseq}
\end{figure}

As shown in \figref{fig:trackseq}, through adjusting the parameter $\alpha_p$, measurement sequences with varying noise levels can be generated.
Specifically, the average IoU between ground truths and simulated measurements is 0.8 under $\alpha_p=0.05$ and 0.6 under $\alpha_p=0.4$.

\section{Semantic-Independent KalmanNet} \label{sec4}
\subsection{Semantic-Independent Encoder} \label{sec4.1}

\begin{figure}[t]
     \centering
     \includegraphics[width=\textwidth]{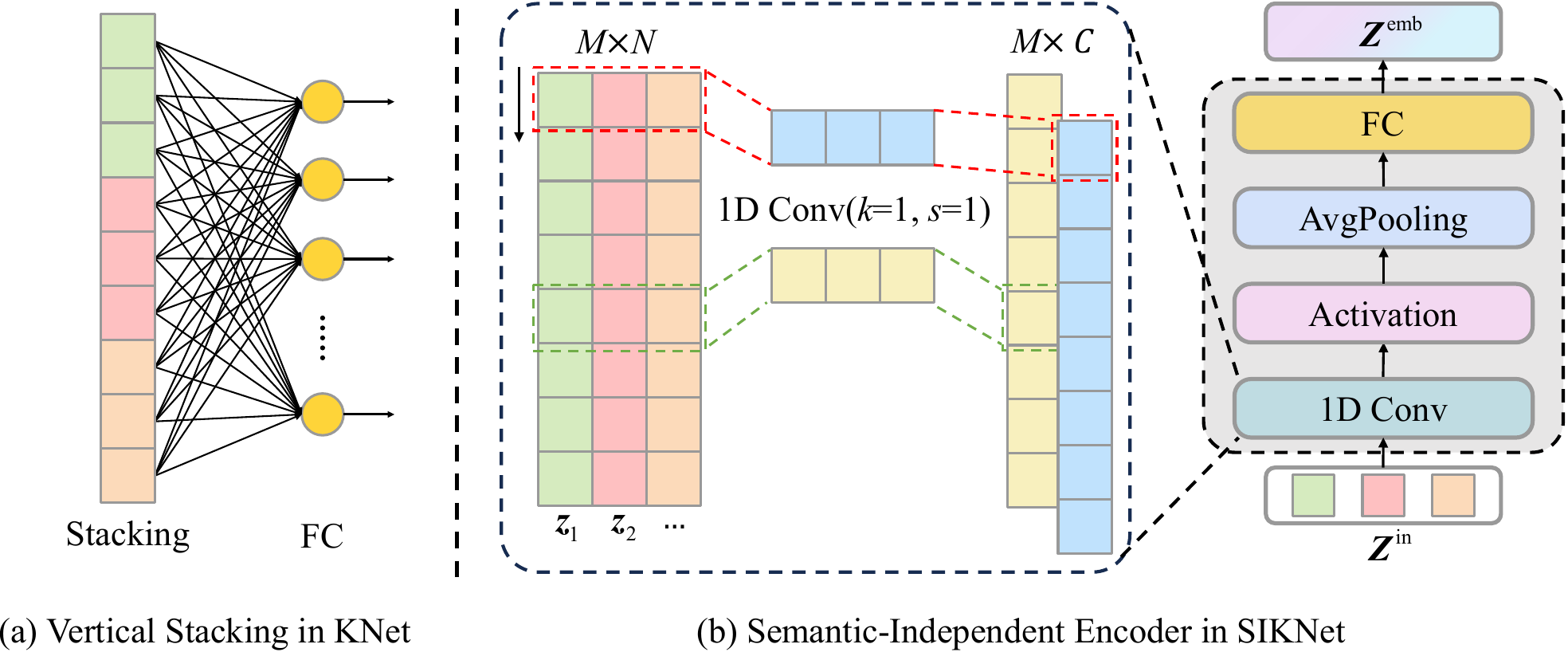}
     \caption{Feature embedding in learning-aided filter. (a) Vertical stacking in KNet. (b) Semantic-Independent Encoder in SIKNet. $\bm{Z}^{\text{in}}$ is composed of $N$ $M$-dimension state (column) vectors $\bm{z}_j \in \mathbb{R}^{M \times 1}$ ($j = 1, 2, \dots, N$). Among them, elements in the same row vector of $\bm{Z}^{\text{in}}=[\bm{z}_1, \bm{z}_2, \dots, \bm{z}_N]$, e.g., $\bm{z}_1 = \bm{\hat{x}}_{t-1|t-1}$ and $\bm{z}_2 = \bm{\hat{x}}_{t|t-1}$, have the same or similar semantic information. By performing a 1D convolution with a convolution kernel of size 1 along the element (row) dimension, we can ensure that state elements with similar semantics are embedded independently between different state vectors. The FC is used to extract the dependencies between elements with different semantic information.}
     \label{fig:SIE}
\end{figure}

While embedding RGB channels into high-dimensional vectors for semantic fusion is common in computer vision \cite{2024donghaomoderntcnmodernpure}, direct concatenation and embedding of state vectors in learning-aided filters is problematic. 
State elements are much more heterogeneous than RGB channels, having different semantic information and numerical scales.
For instance, the state elements $[cx, cy]$ and $[a]$ in Eq. (\ref{eq:state}) have weakly correlation due to their distinct physical meanings.
As shown in \figref{fig:SIE}(a), existing learning-aided filters directly concatenate state vectors with different semantic information (e.g., position and ratio) as input features, which are then processed linearly by a fully-connected (FC) layer. 
Such a way of embedding learning is inappropriate and counterintuitive, because it neglects inherent semantic differences of the input features and undermines the semantic-independence of the heterogeneous state elements, which leads to a degrading quality of feature representation. 
Additionally, given input features with large numerical difference between heterogeneous-semantic elements, the FC layer as well as the subsequent RNN may become instable in network training.
To address this problem, we propose a Semantic-Independent Encoder (SIE), which can be simply expressed as 
\begin{equation} \label{eq7}
    \bm{Z}^{\text{emb}}=\text{Encoder}(\bm{Z}^{\text{in}}),
\end{equation}
where $\bm{Z}^{\text{in}} \in \mathbb{R}^{M \times N}$ is a feature matrix composed of $N$ $M$-dimensional state (column) vectors $\bm{z}_j \in \mathbb{R}^{M\times 1}$ ($j = 1, 2, \dots, N$). 
Crucially, elements in the same row vector of $\bm{Z}^{\text{in}}=[\bm{z}_1, \bm{z}_2, \dots, \bm{z}_N]$, e.g., $\bm{z}_1 = {\bm{\hat{x}}}_{t-1|t-1}$ and $\bm{z}_2 = \bm{\hat{x}}_{t|t-1}$, have the same or similar semantic information.

As shown in \figref{fig:SIE}(b), the SIE consists of a 1-Dimensional (1D) convolutional layer, a non-linear activation function, an adaptive pooling layer, and an FC. 
First, the 1D convolution with a $1 \times N$ kernel is used to extract features from homogeneous-semantic elements along the element (row) dimension. 
Be aware such a 1D convolution is different from the Temporal Convolutional Network \cite{2018kangtcnntubeletsconvolutional,2018baiempiricalevaluationgeneric,2024donghaomoderntcnmodernpure}, which performs convolution along the time dimension. 
Next, the 1D convolution is followed by an activation function, which enables the network to capture non-linearity, e.g., the non-linear transformation from  $\bm{\hat{x}}_{t - 1|t - 1}$ to $\bm{\hat{x}}_{t|t - 1}$. 
The Tanh activation function is selected to converse the input features into quantities within $[-1, 1]$, which help improve the robustness of the subsequent RNN in training and inference. 
Subsequently, the adaptive pooling layer is used to make $\bm{Z}^{\mathrm{emb}}$ an $M$-dimension vector, which can adapt to the input of the subsequent FC. 
Finally, the FC is designed to merge information across heterogeneous-semantic elements. 
SIE can provide higher quality feature representations by decoupling the feature embedding process of homogeneous and heterogeneous semantic elements.

\subsection{Input features}
The SIKNet uses the following features to learn $\bm{{P}}_{t|t-1}$ and $\bm{S}_t^{-1}$:
\begin{enumerate}
    \item[$\bm{Z}^{\text{in}}_1$] \textit{State evolution and update differences} $ [\Delta\bm{\tilde{x}}_t, \Delta\bm{\hat{x}}_t]$. The available feature at time $t$ is $[\Delta\bm{\tilde{x}}_{t-1}, \Delta\bm{\hat{x}}_{t-1}]$.
    \item[$\bm{Z}^{\text{in}}_2$] \textit{Measurement and innovation differences} $[\Delta\bm{\tilde{y}}_t, \Delta\bm{y}_t]$.
    \item[$\bm{Z}^{\text{in}}_3$] \textit{State estimation and its prediction} $[\bm{\hat{x}}_{t-1|t-1}, \bm{\hat{x}}_{t|t-1}, \bm{\hat{x}}_{t|t}]$. 
    The accessible feature at time step $t$ is $[\bm{\hat{x}}_{t-2|t-2},\bm{\hat{x}}_{t-1|t-2}, \\  \bm{\hat{x}}_{t-1|t-1}]$.
    \item[$\bm{Z}^{\text{in}}_4$] \textit{Measurement and its prediction} $[\bm{{y}}_{t-1}, \bm{\hat{y}}_{t|t-1}, \bm{{y}}_{t}]$. 
\end{enumerate}

The first two groups of difference features $\bm{Z}^{\text{in}}_1$ and $\bm{Z}^{\text{in}}_2$, commonly used in existing KalmanNet methods, mainly contribute to the learning of $\bm{Q}_{t}$ and $\bm{R}_{t}$, hence $\bm{{P}}_{t|t-1}$ and $\bm{S}_t^{-1}$;
whereas the second two groups of original features $\bm{Z}^{\text{in}}_3$ and $\bm{Z}^{\text{in}}_4$, purposely selected in our SIKNet, could help learn $\bm{{P}}_{t|t-1}$ and $\bm{S}_t^{-1}$ since them explicitly depend on $\bm{\hat{x}}_{t-1|t-1}$ and $\bm{\hat{x}}_{t|t-1}$, as described in Eqs. (\ref{eq:kfpred}) and (\ref{eq:kfupdate1})-(\ref{eq:kfupdate2}).
Note that the structure design of SIE makes it possible that original features $\bm{Z}^{\text{in}}_3$ and $\bm{Z}^{\text{in}}_4$ could be used in the SIKNet with performance improvement. 
Meanwhile, difference features $\bm{Z}^{\text{in}}_1$ and $\bm{Z}^{\text{in}}_2$ are reserved in the SIKNet to help its training.

\subsection{Overall architecture}
\begin{figure}[!t]
     \centering
     \includegraphics[width=1\textwidth]{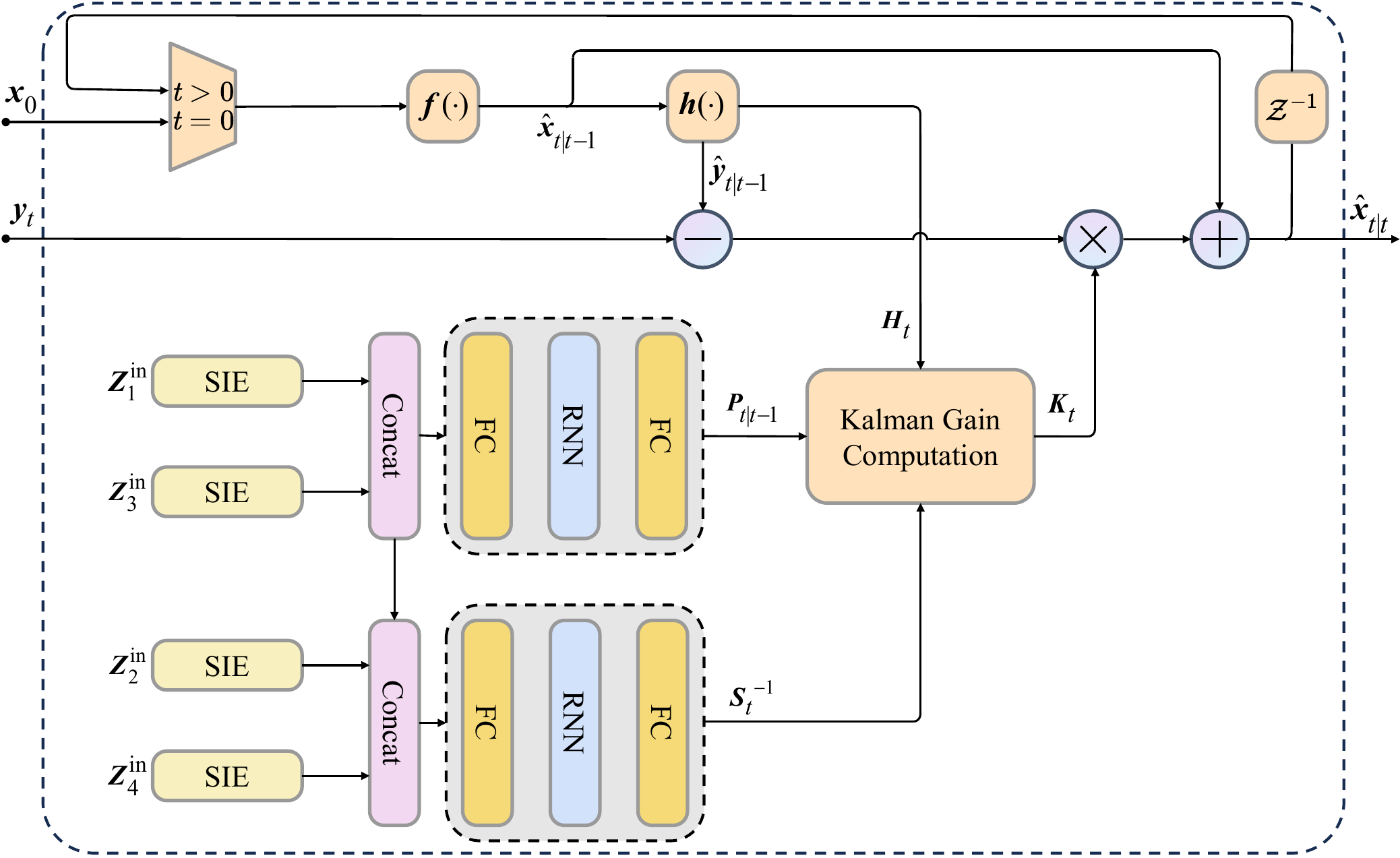}
     \caption{Block diagram of the proposed Semantic-Independent KalmanNet}
     \label{fig:Block}
 \end{figure}
We adopt the similar network structure as SKNet, which has exhibited its training and inference robustness on both simulation and real-world data\cite{2023choisplitkalmannetrobustmodelbased,2024songpracticalimplementationkalmannet}.
As shown in \figref{fig:Block}, four SIE modules are employed by SIKNet to generate embeddings from $\bm{Z}^{\text{in}}_1$, $\bm{Z}^{\text{in}}_2$, $\bm{Z}^{\text{in}}_3$ and $\bm{Z}^{\text{in}}_4$, respectively:
\begin{equation}
    \bm{Z}^{\mathrm{emb}}_{i} = \mathrm{SIE}_i(\bm{Z}^{in}_i),\quad i=1,\dots,4.
    \label{eq:sieprocessing}
\end{equation}
The processed features are fed into two independent DNN modules, each of which integrates FC and RNN to track the $\bm{{P}}_{t|t-1}$ and $\bm{S}_t^{-1}$,
\begin{align}
    \bm{{P}}_{t|t-1} &= \mathcal{G}_t^1\left(\{\bm{Z}^{\mathrm{emb}}_1, \bm{Z}^{\mathrm{emb}}_3\}\right) \label{eq:covdnn}, \\
    \bm{S}_t^{-1} &= \mathcal{G}_t^2\left(\{\bm{Z}^{\mathrm{emb}}_1,\bm{Z}^{\mathrm{emb}}_2,\bm{Z}^{\mathrm{emb}}_3, \bm{Z}^{\mathrm{emb}}_4\}\right). \label{eq:infodnn}
\end{align}
Then, the $\bm{K}_t$ can be computed by:
\begin{equation}
    \bm{K}_t(\theta_1,\theta_2,\bm{H})=\mathcal{G}_t^1{(\theta_1)}\bm{H}^\top\mathcal{G}_t^2{(\theta_2)},
\end{equation}
where $\mathcal{G}_t^1{(\theta_1)}$ and $\mathcal{G}_t^2{(\theta_2)}$ represent two DNN modules with trainable parameters $\theta_1$ and $\theta_2$, respectively.

\subsection{Loss function}
Consider a semi-simulated dataset $\mathcal{D}=\{(\bm{y}_i^{b},\bm{x}_i^{b})\}_{i = 1}^N$ composed of $N$ trajectories with different lengths $T^{i}$, where 
$\bm{y}_i^{b}=[\bm{y}_{1}^{b,{(i)}},\bm{y}_{2}^{b,{(i)}}, \ldots,\bm{y}_{T^{i}}^{b,{(i)}}]$, $\bm{x}_i^{b}=[\bm{x}_{1}^{b,{(i)}},\bm{x}_{2}^{b,{(i)}}, \ldots,\bm{x}_{T^{i}}^{b,{(i)}}]$.
The Smooth L1 Loss \cite{2015girshickfastrcnn} is adopted, with the loss for the $i$-th trajectory defined as:
\begin{equation}
    \mathcal{L}(\theta)_{T^i}=\frac{1}{T^{i}}\sum_{t=1}^{T^i}{\ell_{t}^{i}(\theta)}
\end{equation}
with
\begin{equation}
    \ell_{t}^{i}(\theta) \!=\! 
\begin{cases}
    0.5(\bm{x}_{t}^{b,{(i)}} - {\bm{\hat{x}}_{{t|t}}}^{b,{(i)}})^2 & |\bm{x}_{t}^{b,{(i)}} - \bm{\hat{x}}_{{t|t}}^{b,{(i)}}| < 1 \\
    |\bm{x}_{t}^{b,{(i)}} - \bm{\hat{x}}_{{t|t}}^{b,{(i)}}| - 0.5 & \text{otherwise} .
\end{cases}
\end{equation}

In contrast to the Mean-Square-Error (MSE) loss adopted by KalmanNet, our SIKNet with the Smooth L1 Loss exhibits superior robustness and can better handle the gradient explosion problem caused by outliers.

\section{Experiments} \label{sec5}
% In this section, we conduct an in-depth evaluation of our algorithm against baseline algorithms such as KF, KNet, and SKNet in terms of accuracy and robustness, using the semi-simulated datasets with various noise setups.

\subsection{Simulation experiments}
In this section, we conduct an in-depth evaluation of our algorithm against baseline algorithms (e.g., KF, KNet, SKNet) in terms of motion estimation accuracy and motion estimation robustness, using semi-simulated datasets with various noise setups.

\subsubsection{Experiments setup}

\textbf{Evaluation metrics}: 
Instead of conventional evaluation metrics of root mean square error (RMSE), the recall (Re) defined as below is used for statistic evaluation of the motion estimation:
\begin{equation}
    \mathrm{Re}_{\beta}=\frac{\mathrm{TP}}{\mathrm{TP}+\mathrm{FN}},
\end{equation}
where $\beta$ is the IoU threshold, $\mathrm{TP}$ is the number of true positives, and $\mathrm{FN}$ is the number of false negatives.
An estimated bbox is classified as a TP if its IoU with the corresponding GT bbox meets or exceeds $\beta$.
Otherwise, it is an FN.
Note that for our task of motion estimation there is no case of falsely associating an estimated bbox with the GT box (i.e., false positive), therefore Re instead of F1-score is used here.

To comprehensively evaluate recall across varying IoU thresholds, the average recall (AR) is computed, which is the integral of recall over the interval $[0.5, 1]$, following the convention in object detection \cite{2016hosangwhatmakeseffective}.
This integral, which is equivalent to the area under the IoU-Recall curve and represents the overall performance of the recall metric across different IoU thresholds, is approximated by evaluating recall at 10 equidistant thresholds ($\beta=0.5,0.55,\cdots,0.95$) and averaging the results:
\begin{equation}
    \text{AR}=2\int_{0.5}^1\mathrm{Re}_{\beta}\mathrm{d}\beta\approx\frac{1}{10}\sum_{\beta\in\{0.5, 0.55, \cdots, 0.95\}}\mathrm{Re}_{\beta},
\end{equation}

We use $\text{Re}_{50}$, $\text{Re}_{75}$, and AR to examine estimation performance.
Note that in this section we are only concerned with estimator performance, so we use $\text{Re}{50}$, $\text{Re}{75}$, and AR to examine estimation performance.
MOT-related performance metrics, such as HOTA, MOTA, IDF1, etc., will be demonstrated in subsequent integration experiments.

\textbf{Datasets:}
The $\alpha_p$ for measurement generation as in Eq. (\ref{eq:noise track}) was set to be 0.05, 0.1, 0.2 and 0.4, with $\alpha_p = 0.05$ corresponding to the default setting for most MOT algorithms using the Kalman filter \cite{2016bewleysimpleonlinerealtime,2017wojkesimpleonlinerealtime,2022aharonbotsortrobustassociations,2022zhangbytetrackmultiobjecttracking,2023dustrongsortmakedeepsort}.
Each semi-simulated dataset was divided into training and test sets by splitting each sequence into two equal temporal segments.
And 10\% of training sets sequences were selected as validation sets.
Our main results are reported on the test sets.

\textbf{Compared models:                                                                                         }
\begin{itemize}
    \item KF: The matrices $\bm{F}$, $\bm{H}$, $\bm{Q}$ and $\bm{R}$ follow the setting described in \subsecref{subsec2.1}. The $\alpha_p$ in KF is set to equal the $\alpha_p$ of the test sets and unless stated otherwise. The $\alpha_v$ parameter in the KF's process noise covariance matrix $\bm{Q}$ is set to 0.00625, a commonly adopted value in the literature \cite{2017wojkesimpleonlinerealtime,2016bewleysimpleonlinerealtime,2022zhangbytetrackmultiobjecttracking,2022aharonbotsortrobustassociations}.
    \item KNet and SKNet: Both models were trained using the hyperparameters specified in the original papers \cite{2021revachkalmannetdatadrivenkalman,2022revachkalmannetneuralnetwork, 2023choisplitkalmannetrobustmodelbased}. 
    \item SIKNet: SIKNet is optimized using the Adam optimizer with an initial learning rate of $0.001$, which is gradually decreased to $1 \times 10^{-7}$ using the cosine annealing strategy. The \textit{epoch} and \textit{batch size} are set to 50 and 32, respectively.
\end{itemize}
\textbf{Implementation details:}
Consider the challenges in training KNet, the alternative truncated back-propagation through time algorithm proposed in \cite{2024songpracticalimplementationkalmannet} is employed to ensure successful convergence, which incurs higher training costs.
For SKNet and SIKNet, standard back-propagation through time algorithm is utilized.
All algorithms are implemented in PyTorch on a system featuring an Intel Xeon Platinum 8352V CPU, an NVIDIA GeForce RTX 3080 GPU, and 64 GB RAM.
A modular codebase for reproducibility and comparative analysis is available at \url{https://github.com/SongJgit/filternet}.

\begin{table*}[width=.9\textwidth,cols=4,pos=t]
  \centering
  \caption{Mean recall and mean average recall on test sets. The best results are highlighted in bold and the second-best ones are underlined.}
  \label{tab:alltestset}
  \begin{tabular*}{\tblwidth}{@{}CCCCCCCCCCC@{}}
  \toprule
  \multicolumn{3}{c}{State Mode} & \multicolumn{4}{c}{XYAH} & \multicolumn{4}{c}{XYWH} \\ 
  \cmidrule(lr){1-3}\cmidrule(lr){4-7}\cmidrule(lr){8-11}
  $\alpha_p$ & Metric & Observation & KF & KNet & SKNet & Ours & KF & KNet & SKNet & Ours \\ 
  \toprule
  \multirow{3}{*}{0.05} & $\text{mAR}$ & 0.4872 & 0.6250 & \underline{0.6976} & 0.6389 & \textbf{0.7171} & 0.6067 & \underline{0.7067} & 0.6836 & \textbf{0.7266} \\
   & $\text{mRe}_{50}$ & 0.9509 & \underline{0.9807} & 0.9791 & 0.9464 & \textbf{0.9850} & 0.9831 & \underline{0.9850} & 0.9672 & \textbf{0.9871} \\
   & $\text{mRe}_{75}$ & 0.4248 & 0.7011 & \underline{0.8291} & 0.7465 & \textbf{0.8484} & 0.6603 & \underline{0.8385} & 0.7990 & \textbf{0.8603} \\ 
  \midrule
  \multirow{3}{*}{0.1} & $\text{mAR}$ & 0.2852 & 0.4568 & 0.5720 & \underline{0.5733} & \textbf{0.6154} & 0.4497 & 0.5911 & \underline{0.5979} & \textbf{0.6228} \\
   & $\text{mRe}_{50}$ & 0.7803 & 0.9393 & \underline{0.9646} & 0.9501 & \textbf{0.9720} & 0.9417 & \underline{0.9660} & 0.9596 & \textbf{0.9768} \\
   & $\text{mRe}_{75}$ & 0.1487 & 0.3678 & 0.6001 & \underline{0.6149} & \textbf{0.6748} & 0.3470 & 0.6366 & \underline{0.6501} & \textbf{0.6902} \\ 
  \midrule
  \multirow{3}{*}{0.2} & $\text{mAR}$ & 0.1928 & 0.3799 & \underline{0.5546} & 0.5467 & \textbf{0.6035} & 0.3876 & 0.5635 & \underline{0.5772} & \textbf{0.6043} \\
   & $\text{mRe}_{50}$ & 0.6879 & 0.9370 & \textbf{0.9716} & 0.9311 & \underline{0.9700} & \underline{0.9612} & 0.9457 & 0.9488 & \textbf{0.9741} \\
   & $\text{mRe}_{75}$ & 0.0592 & 0.2114 & 0.5594 & \underline{0.5694} & \textbf{0.6649} & 0.2083 & 0.5977 & \underline{0.6198} & \textbf{0.6585} \\ 
  \midrule
  \multirow{3}{*}{0.4} & $\text{mAR}$ & 0.1465 & 0.2996 & \underline{0.5420} & 0.4888 & \textbf{0.5864} & 0.3307 & 0.5373 & \underline{0.5714} & \textbf{0.5850} \\
   & $\text{mRe}_{50}$ & 0.5846 & 0.8460 & \underline{0.9487} & 0.8526 & \textbf{0.9622} & 0.9239 & 0.9385 & \underline{0.9587} & \textbf{0.9679} \\
   & $\text{mRe}_{75}$ & 0.0344 & 0.1303 & \underline{0.5490} & 0.5026 & \textbf{0.6265} & 0.1407 & 0.5451 & \underline{0.5926} & \textbf{0.6232} \\ 
  \toprule
  % \multicolumn{9}{l}{} & \multicolumn{1}{l}{} & \multicolumn{1}{l}{}
  \end{tabular*}
  \end{table*}

\begin{figure}[h]
    \centering
    \includegraphics[width=0.5\textwidth]{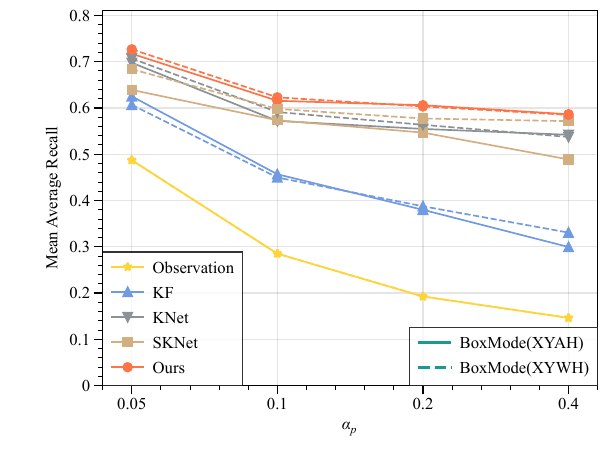}
    \caption{Line graph of mean average recall corresponding to \protect Table \tabref{tab:alltestset}. The x-axis is noise level $\alpha_p$, and the y-axis is the mean average recall.
}
    \label{fig:papersallar}
\end{figure}
\subsubsection{Results on all test sets}\label{subsec: allresults}
Experimental results on all test sets are presented in Table \tabref{tab:alltestset}, which gives the mean recall (mRe) and mean average recall (mAR) of all categories.
The learning-aided filters significantly outperforms the model-based KF in almost all cases, achieving a 40\% average improvement in mAR.
Furthermore, this advantage becomes more pronounced under higher noise levels.
The proposed SIKNet achieves superior SOTA performance compared to existing learning-aided filters, with improvements of approximately 6\% in mAR and 10\% in $\text{mRe}_{75}$.
This demonstrates the ability of SIKNet to provide more accurate bounding-box estimates.

\figref{fig:papersallar} gives the visualized results of Table \tabref{tab:alltestset} for a better observation of the mAR fluctuation. 
For different noise levels $\alpha_p$, SIKNet has the best consistent performance over the two state modes compared to the existing learning-aided filters, which can be attributed to the SIE's capability of decoupling the feature-embedding processes of homogeneous and heterogeneous semantic state elements and effectively reducing sensitivity to differences in state elements.

\subsubsection{Results on different categories}
\begin{figure}[t]
    \centering
    \subfloat[$\alpha_p =0.2$]{%
        \includegraphics[width=0.48\textwidth]{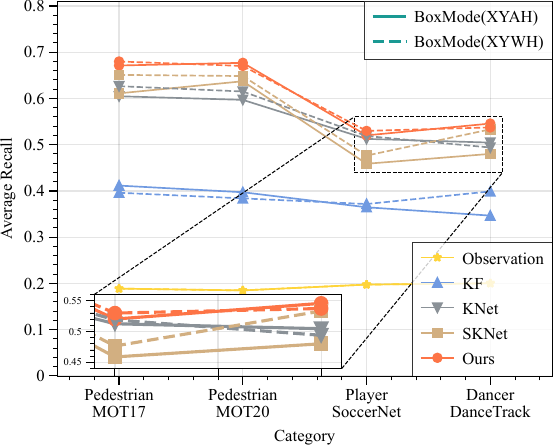}
    }
    \hfill
    \subfloat[$\alpha_p =0.4$]{%
        \includegraphics[width=0.48\textwidth]{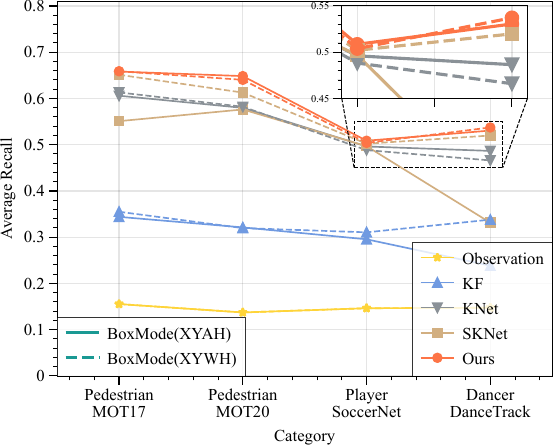}
    }
    \caption{Average recall on different categories. (a) $\alpha_p =0.2$. (b) $\alpha_p =0.4$.} 
    \label{fig:categories}
\end{figure}

Experimental results on different categories at $\alpha_p = 0.2$ are shown in Table \tabref{tab:tab2}, which reveals similarly that the learning-aided methods significantly outperforms the model-based KF in almost all cases, and the proposed SIKNet achieves the SOTA performance in general.
A deeper look at the results suggests that learning-aided filters achieve approximately a 50\% performance improvement on ``Pedestrian'' than the model-based filter, whereas a less improvement of around 35\% on both ``Dancer'' and ``Player''.
This is probably because, as analyzed in \subsecref{subsec3.1}, the motion patterns of ``Dancer'' and ``Player'' change more dramatically than those of ``Pedestrian'' and may cause more serious model-mismatch problem. 
The learning-aided filters could well handle the model-mismatch problem caused by non-linearity, but are less effective for model-mismatch caused by non-stationary.
\figref{fig:categories} visualizes AR fluctuations of each method at $\alpha_p=0.2$ and 0.4 across the two state modes for all categories. 
Similarly, SIKNet demonstrates the highest consistency over the two state modes.
 \begin{table}[t]
      \caption{Recall and average recall for different categories on the test sets using $\alpha_p=0.2$.}
      \centering
      \label{tab:tab2}
        \begin{tabular*}{\tblwidth}{@{}CCCCCCCCCCCC@{}}
          \toprule
          \multicolumn{4}{c}{State Mode} & \multicolumn{4}{c}{XYAH} & \multicolumn{4}{c}{XYWH} \\ 
          \cmidrule(lr){1-4}\cmidrule(lr){5-8}\cmidrule(lr){9-12}
          \multicolumn{1}{c}{Datasets} & \multicolumn{1}{c}{Category} & \multicolumn{1}{c}{Metric} & \multicolumn{1}{c}{Obervation} & KF & KNet & SKNet & \multicolumn{1}{c}{Ours} & KF & KNet & SKNet & Ours \\ 
          \toprule
          \multirow{3}{*}{MOT17} & \multirow{3}{*}{Pedestrian} & $\text{AR}$ & 0.1887 & 0.4115 & 0.6047 & \underline{0.6110} & \textbf{0.6712} & 0.3960 & 0.6265 & \underline{0.6509} & \textbf{0.6792} \\
           &  & $\text{Re}_{50}$ & 0.7055 & 0.9620 & \textbf{0.9932} & 0.9581 & \underline{0.9771} & 0.9670 & 0.9579 & \underline{0.9679} & \textbf{0.9918} \\
           &  & $\text{Re}_{75}$ & 0.0509 & 0.2462 & 0.6503 & 0.6860 & \textbf{0.7990} & 0.2107 & 0.7286 & \underline{0.7603} & \textbf{0.7994} \\ 
          \midrule
          \multirow{3}{*}{MOT20} & \multirow{3}{*}{Pedestrian} & $\text{AR}$ & 0.1848 & 0.3971 & 0.5969 & \underline{0.6371} & \textbf{0.6770} & 0.3838 & 0.6150 & \underline{0.6484} & \textbf{0.6702} \\
           &  & $\text{Re}_{50}$ & 0.6573 & 0.9557 & \underline{0.9826} & 0.9791 & \textbf{0.9838} & 0.9590 & \underline{0.9741} & 0.9537 & \textbf{0.9844} \\
           &  & $\text{Re}_{75}$ & 0.0563 & 0.2320 & 0.6429 & \underline{0.7250} & \textbf{0.8036} & 0.2053 & 0.6853 & \underline{0.7678} & \textbf{0.7985} \\ 
          \midrule
          \multirow{3}{*}{\begin{tabular}[c]{@{}c@{}}Soccer\\Net\end{tabular}} & \multirow{3}{*}{Player} & $\text{AR}$ & 0.1974 & 0.3646 & \underline{0.5125} & 0.4587 & \textbf{0.5201} & 0.3716 & \underline{0.5187} & 0.4764 & \textbf{0.5297} \\
           &  & $\text{Re}_{50}$ & 0.6931 & 0.9186 & \underline{0.9456} & 0.8529 & \textbf{0.9500} & 0.9361 & \textbf{0.9592} & 0.8947 & \underline{0.9509} \\
           &  & $\text{Re}_{75}$ & 0.0638 & 0.1979 & \underline{0.4912} & 0.4402 & \textbf{0.5151} & 0.1986 & \underline{0.4926} & 0.4467 & \textbf{0.5144} \\ 
          \midrule
          \multirow{3}{*}{\begin{tabular}[c]{@{}c@{}}Dance\\Track\end{tabular}} & \multirow{3}{*}{Dancer} & $\text{AR}$ & 0.2002 & 0.3465 & \underline{0.5045} & 0.4801 & \textbf{0.5456} & 0.3989 & 0.4938 & \underline{0.5331} & \textbf{0.5372} \\
           &  & $\text{Re}_{50}$ & 0.6956 & 0.9115 & \underline{0.9648} & 0.9343 & \textbf{0.9691} & \textbf{0.9826} & 0.8914 & \underline{0.9788} & 0.9693 \\
           &  & $\text{Re}_{75}$ & 0.0659 & 0.1696 & \underline{0.4532} & 0.4263 & \textbf{0.5418} & 0.2188 & 0.4842 & \underline{0.5043} & \textbf{0.5219} \\
           \bottomrule
          \end{tabular*}
  \end{table}

\subsubsection{Results on mismatched noise conditions}
\begin{figure*}[!t]
    \centering
    \subfloat[Trained on $\alpha_p= 0.05$]{\includegraphics[width=0.48\textwidth]{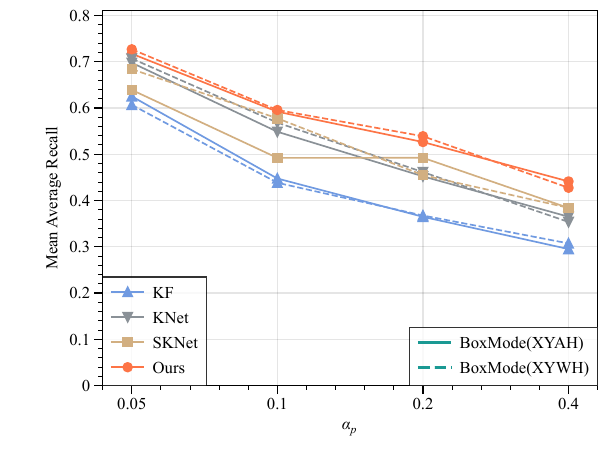}%
    }
    \hfil
    \subfloat[Trained on $\alpha_p= 0.4$]{\includegraphics[width=0.48\textwidth]{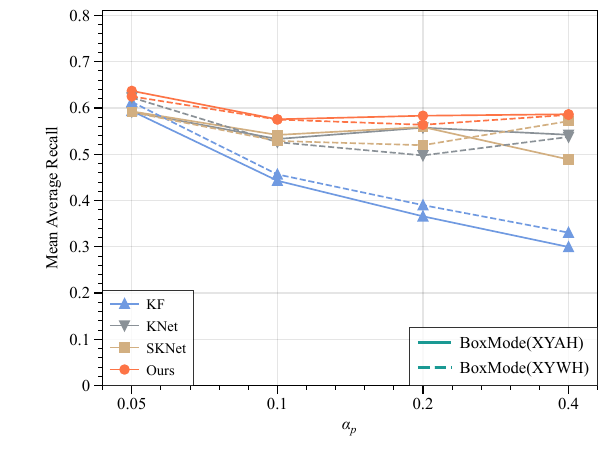}%
    }
    \caption{Mean average recall for scenarios where training and testing noise parameters are mismatched. (a) Trained on $\alpha_p = 0.05$. (b) Trained on $\alpha_p = 0.4$.}
    \label{fig:mismatch}
\end{figure*}
Experimental results visualized in \figref{fig:mismatch} are for scenarios where training and testing noise parameters are mismatched. 
The learning-aided methods were trained on datasets generated with $\alpha_p = 0.05$ and 0.4, respectively, and tested on datasets with $\alpha_p \in \{0.05, 0.1, 0.2, 0.4\}$.
The model-based KF with parameters $\alpha_p = 0.05$ or 0.4 was as well tested on datasets with $\alpha_p$ = 0.05, 0.1, 0.2 and 0.4, respectively. 
As we can see, for all situations when the noise parameters are mismatched, the proposed SIKNet remains the best and all the learning-aided methods are better than the model-based KF. 

\begin{figure*}[t]
     \centering
     \includegraphics[width=\textwidth]{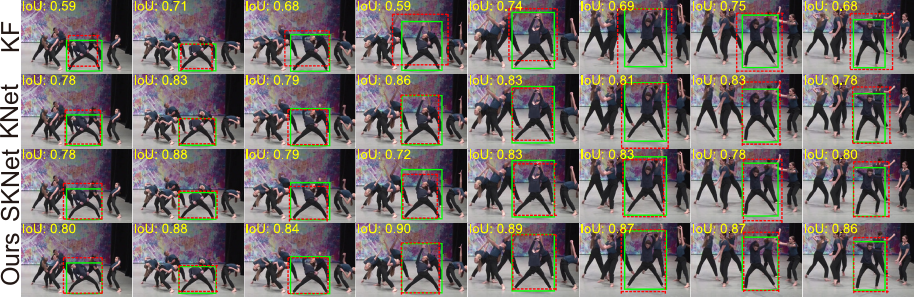}
     \caption{
         Visual results of selected sequences obtained by different motion estimation methods on the DanceTrack in \textit{XYAH} mode.
         Training and test sets of the sequences are generated using $\alpha_p = 0.2$. The frame interval of the sequences is 5 and the images were cropped to focus on the main track area. Ground-truth bounding-boxes are marked by green solid lines and estimated results by red dashed lines.}
     \label{fig:predtrack}
   \end{figure*}
\subsubsection{Selected visual results}
Visual results of selected sequences on the DanceTrack datasets are presented in \figref{fig:predtrack}.
The sequence is taken from the last few frames of a 600-frame video, where the object has moved continuously for 600 frames.
The results show that SIKNet achieves the highest average IoU (0.86), outperforming KNet (0.81), SKNet (0.79), and KF (0.68). Additionally, SIKNet demonstrates the most stable performance with an IoU variance of $0.8 \times 10^{-3}$ on this sequence, followed by KNet ($1.0 \times 10^{-3}$), SKNet ($1.2 \times 10^{-3}$), and KF ($3.8 \times 10^{-3}$). 
These results highlight the efficacy of the learning-aided filter in processing long-term sequences.

\subsection{Performance comparison of trackers integrated with learning-aided kalman filtering}
In this section, we integrate several LAKF models into existing trackers—replacing the model-based (MB) Kalman Filter (KF)—to validate the improvements that LAKF models bring to the trackers.

\textbf{Evaluation Metrics:} To evaluate the tracking results comprehensively across different dimensions, we employ HOTA\cite{2021luitenhotahigherorder}, AssA, MOTA, IDF1, and ID switches (IDs) \cite{2008bernardinevaluatingmultipleobject}.

\textbf{Datasets:}
We conducted experiments on two MOT datasets where irregular motion is most prominent: DanceTrack \cite{2022sundancetrackmultiobjecttracking} and SoccerNet \cite{2022cioppasoccernettrackingmultipleobject}. 

\textbf{Implementation details:} 
We focus on the impact of different motion models on trackers, so we used the famous BYTE (using its original parameters) as the association algorithm for all experiments. 
BYTE + KF is the well-known ByteTrack.
All LAKF models are first trained on semi-simulated datasets generated from the training sets of DanceTrack and SoccerNet, where $\alpha_{p}=0.05$.
They are then combined with BYTE and evaluated separately on the validation set of DanceTrack \cite{2022sundancetrackmultiobjecttracking} and the test set of SoccerNet \cite{2022cioppasoccernettrackingmultipleobject}.
DanceTrack and SoccerNet use oracle detections.

\begin{table}
\centering
\caption{Comparison of different motion models on DanceTrack, SoccerNet. The association algorithm is BYTE.}
\label{tab:mot}
\begin{tabular}{c|l|ccccc} 
\toprule
Datasets                    & Motion Model & HOTA$\uparrow$ & AssA$\uparrow$ & MOTA$\uparrow$ & IDF1$\uparrow$ & IDs$\downarrow$  \\ 
\midrule
\multirow{4}{*}{DanceTrack} & KF           & 49.95          & 34.80          & 90.41          & 56.22          & 1738             \\
                            & KNet         & \underline{54.60}  & \underline{38.88}  & \underline{92.10}  & \textbf{58.20} & \underline{1483}     \\
                            & SKNet        & 50.94          & 35.74          & 89.99          & 54.88          & 1619             \\
                            & SIKNet(Ours) & \textbf{56.19} & \textbf{39.97} & \textbf{92.37} & \underline{57.87}          & \textbf{1427}    \\ 
\midrule
\multirow{4}{*}{SoccerNet}  & KF           & 72.30          & 62.48          & 94.62          & 75.58          & 5054             \\
                            & KNet         & \underline{75.49}  & \underline{66.67}          & \underline{94.71}           & \underline{77.71}          & 4354              \\
                            & SKNet        & 73.82          & 65.14          & 94.24          & 77.43          & \underline{4171}             \\
                            & SIKNet(Ours) & \textbf{76.17} & \textbf{67.45} & \textbf{95.43} & \textbf{77.81} & \textbf{3844}    \\ 
\bottomrule
\multicolumn{7}{l}{Best results are shown in bold, second best underlined.}                                                       \\
\end{tabular}
\end{table}
\subsubsection{Results}
The experimental results are reported in Table \ref{tab:mot}.
As indicated by the results, LAKF models—SIKNet in particular, which outperforms other LAKF models across nearly all metrics—surpass KF by a large margin across all metrics on the two datasets.

\section{Conclusions} \label{sec6}
In this paper, we have proposed a novel learning-aided filter SIKNet featuring a Semantic-Independent Encoder (SIE) for robust motion estimation. 
The SIE utilizes a 1D convolutional layer and a fully connected layer to decouple the feature-embedding processes of homogeneous and heterogeneous semantic elements in the state vector. 
It effectively mitigates the training robustness problem of KNets caused by the heterogeneous-semantic and numerical range differences of the state elements. 
To validate the proposed algorithm, we develop a semi-simulated dataset based on several well-known MOT real datasets. 
Experimental results with various noise levels show that SIKNet outperforms the model-based KF and existing learning-aided filters in terms of both accuracy and robustness. 
Specifically, SIKNet achieves a 6\% improvement in mAR over existing learning-aided filters and a 40\% improvement over the model-based KF. 
Due to joint optimization challenges, SIKNet has not yet been fully integrated into complete MOT frameworks for end-to-end training with detection components. Future work will focus on seamless integration of learning-aided filter into full MOT pipelines to implement joint optimization and enhance overall tracking performance.

\bibliographystyle{elsarticle-num}
% \bibliographystyle{elsarticle-harv}
\bibliography{reference}



\end{document}